\documentclass[10pt,conference]{IEEEtran}
\usepackage{braket,amsfonts}
\usepackage{array}
\usepackage{subcaption}
\usepackage{pgfplots}
\usepackage{algorithmic}
\usepackage{graphicx,epstopdf}
\usepackage{amsopn}

\usepackage{amsthm}
\newtheorem{theorem}{Theorem}

\usepackage{amsmath}
\usepackage[utf8]{inputenc} 
\usepackage[T1]{fontenc}    
\usepackage{hyperref}       
\usepackage{url}            
\usepackage{booktabs} 
\usepackage{amsfonts}       
\usepackage{nicefrac}       
\usepackage{microtype}      
\usepackage{mathrsfs}
\usepackage{textcomp}
\usepackage{epsfig}
\usepackage{amssymb}
\usepackage{amsmath}
\usepackage{amsfonts}
\usepackage{algorithmic}
\usepackage{algorithm}
\usepackage{multirow}
\usepackage{color}
\usepackage{wrapfig}
\usepackage{pgfplots}
\usepackage{graphicx}
\usepackage{mathtools}
\usepackage{bbm}
\usepackage[english]{babel}
\usepackage{lipsum}
\usepackage{multicol}
\usepackage{todonotes} 
 
\usepackage{xspace}
\usepackage{bold-extra}
\usepackage[most]{tcolorbox}
\newtheorem{lemma}[theorem]{Lemma}
\usepackage{booktabs}
\usepackage{mathrsfs}
\usepackage{epsfig}
\usepackage{amssymb}
\usepackage{amsmath}
\usepackage{amsfonts}
\usepackage{algorithmic}
\usepackage{algorithm}
\usepackage{multirow}
\usepackage{cite}
\usepackage{mathtools,bbm}
\usepackage{enumerate}
\usepackage{lipsum}
\usepackage{mathtools}
\usepackage{cuted}
\usepackage{amsopn}

\usepackage{cancel}
\usepackage{verbatim} 
\usepackage{mathtools}
\usepackage{amssymb}
\usepackage{amsmath}
\usepackage{amsfonts}
\usepackage[english]{babel}

\usepackage[normalem]{ulem}
\usepackage{hyperref}

\makeatletter
\DeclareUrlCommand\ULurl@@{%
  \def\UrlLeft{\uline\bgroup}%
  \def\UrlRight{\egroup}}
\def\ULurl@#1{\hyper@linkurl{\ULurl@@{#1}}{#1}}
\DeclareRobustCommand*\ULurl{\hyper@normalise\ULurl@}
\makeatother

\def\BibTeX{{\rm B\kern-.05em{\sc i\kern-.025em b}\kern-.08em
    T\kern-.1667em\lower.7ex\hbox{E}\kern-.125emX}}

 \usepackage{scalerel,amssymb}

\usepackage[utf8]{inputenc}
\usepackage[english]{babel}
\usepackage{amsthm}
\usepackage{mathtools}

\newtheorem{definition}{Definition}[section]

\usepackage{tabularx}

\usepackage{algorithm}
\usepackage{algorithmic}

\usepackage{adjustbox}

\usepackage[normalem]{ulem}
\usepackage{hyperref}

\begin{document}
 
\title{HyperEF: Spectral Hypergraph Coarsening by  Effective-Resistance Clustering}

\author{\IEEEauthorblockN{ Ali Aghdaei }
\IEEEauthorblockA{
{Stevens Institute of Technology}\\
aaghdae1@stevens.edu }

\and
\IEEEauthorblockN{Zhuo Feng}
\IEEEauthorblockA{
{Stevens Institute of Technology}\\
zhuo.feng@stevens.edu}

 }

\maketitle

\begin{abstract}
This paper introduces a scalable algorithmic framework (HyperEF) for spectral coarsening (decomposition) of large-scale hypergraphs by exploiting hyperedge effective resistances. Motivated by the latest theoretical framework for     low-resistance-diameter decomposition of simple graphs, HyperEF aims at decomposing large hypergraphs into multiple node clusters with only a few inter-cluster hyperedges. The key component in HyperEF is a nearly-linear time  algorithm for estimating hyperedge  effective resistances, which allows incorporating the latest diffusion-based non-linear quadratic operators defined on hypergraphs. To achieve good runtime scalability, HyperEF searches within the Krylov subspace (or approximate eigensubspace) for identifying the nearly-optimal vectors for approximating the hyperedge effective resistances. In addition, a  node weight propagation scheme for multilevel spectral hypergraph decomposition   has been introduced for achieving even greater  node coarsening ratios. When compared with state-of-the-art hypergraph partitioning (clustering) methods, extensive experiment results on real-world VLSI designs show that HyperEF can more effectively coarsen (decompose) hypergraphs   without losing  key structural (spectral) properties of the original hypergraphs, while achieving over $70\times$ runtime speedups over hMetis and $20\times$ speedups over HyperSF.
\end{abstract}


\section{Introduction}
Recent years have witnessed a surge of interest in  graph learning  techniques  for various   applications such as vertex (data) classification \cite{perozzi2014deepwalk,grover2016node2vec}, link prediction (recommendation systems) \cite{schlichtkrull2018modeling,zhang2018link},   drug discovery \cite{rathi2019practical,lim2019predicting }, solving partial differential equations (PDEs) \cite{li2020multipole, iakovlev2020learning}, and electronic design automation (EDA) \cite{ zhang2019circuit,kunal2020gana, wang2020gcn,mirhoseini2021graph}. The ever-increasing  complexity of modern graphs (networks)  inevitably demands the development of efficient  techniques  to reduce the size of the input datasets while preserving the essential properties. 
To this end,  many research works on graph partitioning, graph embedding, and graph neural networks (GNNs) exploit graph coarsening techniques to improve the algorithm scalability, and accuracy \cite{safro2015advanced, deng2019graphzoom, zhao2021towards,chen2022graph}.

Hypergraphs are more general  than simple graphs since they  allow modeling  higher-order relationships among the entities \cite{HGreview}.   
The state-of-the-art hypergraph coarsening techniques are all based on relatively simple heuristics, such as the vertex similarity or hyperedge similarity \cite{karypis1999multilevel, devine2006parallel, vastenhouw2005two,ccatalyurek2011patoh,Shaydulin_2019}. For example, the hyperedge similarity based coarsening techniques  contract  similar hyperedges with large sizes into smaller ones, which can be easily implemented but may impact the original hypergraph structural properties; the vertex-similarity based algorithms rely on checking the distances between  vertices for discovering  strongly-coupled (correlated) node  clusters, which can be achieved  leveraging  hypergraph embedding  that maps each vertex into a  low-dimensional vector such that the Euclidean distance (coupling) between the vertices can be easily computed in constant time. However, these \textbf{simple metrics  often fail to capture  higher-order global (structural) relationships in  hypergraphs}.


The latest theoretical breakthroughs in spectral graph theory have led to the development of nearly-linear time spectral graph sparsification (edge reduction) \cite{spielman2011spectral,feng2020grass,Lee:2017,zhuo:dac18,kapralov2022spectral, kapralov2021towards, zhang2020sf} and coarsening (node reduction) algorithms \cite{loukas2018spectrally, zhao:dac19}. However,  \textbf{existing  spectral  methods are only applicable to simple graphs} but not hypergraphs. For example, the effective-resistance based spectral sparsification method \cite{spielman2011graph} exploits an effective-resistance based edge sampling scheme, while the latest practically-efficient sparsification algorithm \cite{feng2020grass} leverages generalized eigenvectors  for   recovering the most influential off-tree edges for mitigating the  mismatches between the subgraph and the original graph. 

On the other hand, spectral theory for hypergraphs has been less developed due to the more complicated structure of hypergraphs. For example, a classical spectral method has been proposed  for hypergraphs by converting each hyperedge into  undirected edges using star or clique expansions \cite{hagen1992new}. However, such a naive hyperedge conversion scheme may result in lower performance due to ignoring the multi-way high-order relationship between the entities. 
 A more rigorous approach by Tasuku and Yuichi \cite{soma2018spectral} generalized  spectral graph sparsification for hypergraph setting by sampling each hyperedge according to a probability determined based on the ratio of the hyperedge weight over the minimum degree of two vertices inside the hyperedge. 
Another family of spectral methods for hypergraphs explicitly builds the Laplacian matrix to analyze the spectral properties of hypergraphs: Zhou et al. proposes a method to create the Laplacian matrix of a hypergraph and generalize graph learning algorithms for hypergraph applications \cite{zhou2006learning}.  A more mathematically rigorous approach by Chan et al. introduces a nonlinear  diffusion process for defining the hypergraph Laplacian operator by measuring the flow distribution within each hyperedge \cite{chan2018spectral,chan2020generalizing}; moreover, the Cheeger's inequality has been proved for  hypergraphs under the diffusion-based nonlinear Laplacian operator \cite{chan2018spectral}. However, these   theoretical results do not immediately allow for  practically-efficient implementations.

This paper presents a scalable spectral hypergraph coarsening algorithm via effective-resistance clustering to generate much smaller hypergraphs that can still well preserve the original hypergraph structural (global) properties.  The key contribution of this work is summarized below:
\begin{enumerate}
\item  To the best of our knowledge, for the first time, we propose to extend the simple-graph  effective resistance  formulation to  the hypergraph setting  by exploiting  approximate eigenvectors (Krylov subspace) associated with the  hypergraph.
\item Our approach (HyperEF)  allows incorporating the recently developed  diffusion-based non-linear Laplacian operator defined on hypergraphs \cite{chan2018spectral} for highly-efficient estimation of  hyperedge effective resistances.
\item A scalable spectral hyperedge contraction (clustering) scheme and a node weight propagation scheme  have been proposed for  constructing coarse-level hypergraphs that can well preserve the  original spectral (structural) properties.  
\item Our extensive experiment results on real-world VLSI designs show that HyperEF is over $70\times$ faster than hMetis, while achieving comparable or better average conductance values for most hypergraph decomposition tasks.
\end{enumerate}

The rest of the paper is organized as follows. In Section  \ref{sec:background}, we provide a background introduction to the basic concepts related to the spectral hypergraph theory. In Section \ref{sec:eff_resistance}, we introduce an optimization-based formulation for effective resistance estimation in simple graphs. In section \ref{sec:spectral_coarsening}, we introduce the  technical details of HyperEF and its algorithm flow. In section \ref{sec:multilevel}, we introduce a multilevel spectral hypergraph decomposition framework. In section \ref{sec:complexity_analysis}, we provide the complexity  analysis of HyperEF. In Section \ref{sec:results}, we demonstrate extensive experimental results to evaluate the performance of HyperEF using a variety of real-word VLSI design benchmarks, which is followed by the conclusion of this work in Section \ref{sec:conclusion}.



\section{Background}\label{sec:background}
Classical spectral graph theory shows that the structure of a simple graph is closely related to the graph's spectral properties. Specifically, the Cheeger's inequality shows the close connection between  expansion or \textit{conductance}  and the first few eigenvalues of graph Laplacians \cite{lee2014multiway}. Moreover, the Laplacian quadratic form computed with the Fiedler vector (the eigenvector corresponding to the smallest nonzero Laplacian eigenvalue) has been exploited to find the minimum boundary size or \textit{cut} for graph partitioning tasks \cite{spielmat1996spectral}. In recent years, spectral theories for hypergraphs have been extensively studied by theoretical computer scientists \cite{chan2018spectral}. To allow a better understanding of our work, the following important definitions    for hypergraphs are introduced.

\begin{table*}[]
\caption{Symbol descriptions}
\begin{adjustbox}{width=1\textwidth}
\label{table:symbols}
\begin{tabular}{|cc|cc|cc|}
\hline
symbols                                     & descriptions                                  & symbols                             & descriptions                                                   & symbols             & descriptions                                 \\ \hline
$H = (V, E, w)$                             & weighted hypergraph                           & $G = (\mathcal{V}, \mathcal{E}, z)$ & weighted simple graph                                          & $H' = (V', E', w')$ & coarsened hypergraph                         \\
$E$                                         & hyperedge set                                 & $\mathcal{E}$                       & simple graph edge set                                          & $V'$                & node set in coarsened hypergraph             \\
$w$                                         & hyperedge weight set                          & $z$                                 & simple graph edge weight set                                   & $E'$                & hyperedge set in coarsened hypergraph        \\
$\mathcal{C}_H(S)$                          & conductance of set $S$ in $H$      & $R_{eff}(p,q)$                      & effective resistance between nodes $p$ and $q$ in $G$ & $w'$                & hyperedge weight set in coarsened hypergraph \\
$\mathcal{C}$                               & average conductance of clusters in $H$ & $r_e^{i}$                           & hyperedge ratio of $e$ associated with $\chi^{(i)}$            & $\rho$              & the order of Krylov subspace                 \\
$Q_H(\chi)$                                 & non-linear quadratic form in $H$       & $R_e$                               & effective resistance of hyperedge $e$                          & $\chi$              & node embedding vector in $H$          \\
$G_b = (\mathcal{V}_b, \mathcal{E}_b, z_b)$ & bipartite graph of $H$                 & $x$                                 & node embedding vector in $G$                          & $d_u$               & number of hyperedges belong to node $u$      \\
$\mathcal{V}_b$                             & node set in bipartite graph                   & $\mathcal{C}_G$                     & conductance of $G$                                       & $L$                 & number of levels coarsening the hypergraph   \\
$\mathcal{E}_b$                             & edge set in bipartite graph                   & $m$                                 & number of resistance ratios               & $\eta$              & node weights in hypergraph                   \\
$z_b$                                       & edge weight set in bipartite graph            & $R$                                 & vector of hyperedge effective resistances                      & $\delta$            & effective resistance threshold               \\ \hline
\end{tabular}
\end{adjustbox}
\end{table*}
\begin{definition}\label{def1}
Let $H = (V, E, w)$ denote a weighted hypergraph, where $w:E \rightarrow \mathbb{R}_+$ is a weight function over hyperedges, $d_u := \Sigma_{u \in e, e \in E}w_e$,  and $vol(S):= \Sigma_{S\in V: u\in S} d_u$. The conductance of a given node set $S \in V$ is defined as
\begin{equation}
    \mathcal{C}_H(S):= \frac{cut(S, \bar{S})}{min\{vol(S), vol(\bar{S})\}}, 
\end{equation}
where the cut is the sum of the weights of the hyperedges containing the nodes from both $S$ and $\bar{S}$. The hypergraph conductance is defined as $\mathcal{C}_H := \min\limits_{\emptyset \nsubseteq S \subseteq V} \mathcal{C}_H(S)$.
\end{definition}

\begin{definition}
The non-linear quadratic form of a hypergraph $H = (V, E, w)$  for any input vector $\chi \in \mathbb{R}^V$ is defined as \cite{chan2018spectral}
\begin{equation}\label{eq:non-linerQ}
    Q_H(\chi) := \sum\limits_{e\in E} w_e \max\limits_{u, v \in e} (\chi_u - \chi_v)^2.
\end{equation}

\end{definition}


\section{Effective Resistances in Simple Graphs}\label{sec:eff_resistance} 

\begin{definition}
Let $G = (\mathcal{V}, \mathcal{E}, z)$ denote a  weighted and connected undirected graph with weights $z \in \mathbb{R}^\mathcal{V}_{\geq 0}$,  ${b_{p}} \in \mathbb{R}^{\mathcal{V}}$ denote  the standard basis vector with all zero entries except for the $p$-th entry being $1$, and    ${b_{pq}}=b_p-b_q$, respectively. The effective resistance between nodes   $(p, q) \in \mathcal{V}$ is defined as
\begin{equation}\label{eq:eff_resist0}
    R_{eff}(p,q) = b_{pq}^\top L_G^{\dagger} b_{pq}=\sum\limits_{i= 2}^{|\mathcal{V}|} \frac{(u_i^\top b_{pq})^2}{\lambda_i}=\sum\limits_{i= 2}^{|\mathcal{V}|} \frac{(u_i^\top b_{pq})^2}{u_i^\top L_G u_i},
\end{equation}
where $L_G^{\dagger}$ denotes the Moore-Penrose pseudo-inverse of the graph Laplacian matrix  $L_G$, and  $u_{i} \in \mathbb{R}^{\mathcal{V}}$ for $i=1,...,|\mathcal{V}|$ denote the  unit-length, mutually-orthogonal  eigenvectors corresponding to  Laplacian eigenvalues $\lambda_i$ for $i=1,...,|\mathcal{V}|$.
\end{definition}
\begin{theorem}
The effective resistance between $p$ and $q$ can be computed by solving the following optimization   problem:
\begin{equation} \label{eq:eff_resist}
    R_{eff}(p,q) = \max\limits_{x \in \mathbb{R}^\mathcal{V}} \frac{(x^\top b_{pq})^2}{x^\top L_G x}.
\end{equation}
\end{theorem}
\begin{proof}
The original problem in (\ref{eq:eff_resist}) can be alternatively solved by tackling the following convex optimization problem:
\begin{equation}
\begin{split}
     \min\limits_{x \in \mathbb{R}^\mathcal{V}} x^\top L_G x \quad\\ s.t. \quad x^\top b_{pq} = 1.
     \end{split}
\end{equation}
After introducing  a Lagrangian multiplier $\lambda$, the following  Lagrangian function needs to be minimized
\begin{equation}
  F(x, \lambda)=  x^\top L_G x -\lambda (x^\top b_{pq}-1), 
\end{equation}
which will lead to the following optimal solution:
\begin{equation}\label{eq:x}
 \lambda^* = \frac{1}{R_{eff}(p,q)}, ~~~x^* = \frac{L_G^{\dagger}b_{pq}}{R_{eff}(p,q)}.
\end{equation}
Substituting $x^*$ into  (\ref{eq:eff_resist}) will  complete the proof.
\end{proof}

\section{Spectral Coarsening via Resistance Clustering}\label{sec:spectral_coarsening}
\begin{figure}
    \centering
    \includegraphics [width = \linewidth]{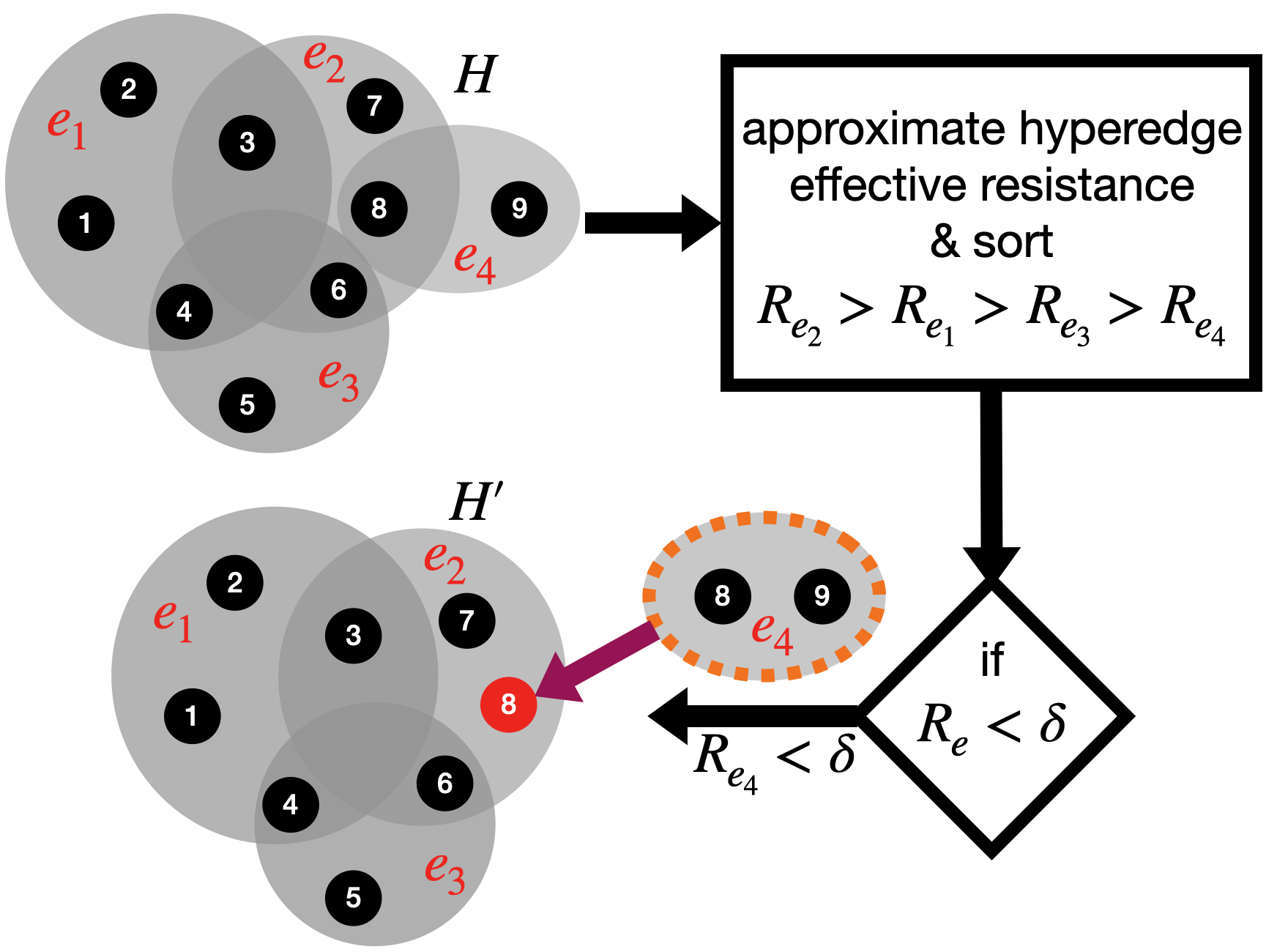}
    \caption{Overview of the proposed HyperEF framework.}
    \label{fig:overview}
\end{figure}
 \subsection{Overview of HyperEF}
 A recent theoretical work  proves that it is possible to decompose a simple graph   into multiple node clusters   with effective-resistance diameter at most the inverse of average node degree (up to constant losses) by removing only a constant fraction of edges \cite{alev2018graph}. This result also leads to the development of algorithms  that use    effective-resistance     metric for spectral graph clustering (decomposition) and spectral graph sparsification \cite{alev2018graph}. 
 
 In \cite{kapralov2022spectral}, an effective-resistance based spectral algorithm has been proposed for    sparsification of hypergraphs. Specifically, by sampling each hyperedge according to its effective resistance, nearly-linear-sized sparsifiers can be obtained for hypergraphs \cite{kapralov2022spectral}. However, such a method involves a non-trivial procedure for estimating hyperedge effective resistances and may not allow for practically efficient implementations since the required hypergraph-to-graph conversion using  clique expansion and the iterative edge weight updating scheme will drastically increase the algorithm complexity \cite{kapralov2022spectral}.
 
 Inspired by the aforementioned research,  we propose a spectral hypergraph coarsening method (HyperEF) that aggressively clusters the nodes within   each hyperedge with low effective-resistance diameters (as shown in Figure \ref{fig:overview}), such that   much smaller   hypergraphs can be constructed without impacting the key structural properties of the original hypergraph. A key technical component in HyperEF is a highly-efficient scheme for estimating hyperedge effective resistances, which is achieved by extending the optimization-based effective-resistance estimation method in   (\ref{eq:eff_resist}) to  the hypergraph setting: hyperedge effective resistance can be obtained  by searching for an optimal vector $\chi^*$ leveraging the following optimization procedure:
\begin{equation} \label{eq:eff_resist_hypergraph}
\begin{split}
 ~R_e &=
\max\limits_{\chi\in \mathbb{R}^\mathcal{V}} \frac{(\chi^{\top} b_{pq})^2}{\sum\limits_{e\in E} w_e \max\limits_{u, v \in e} (\chi_u - \chi_v)^2}, 
\end{split}
\end{equation}
where   the   original quadratic form $x^\top L_G x$ in (\ref{eq:eff_resist})  is replaced by the nonlinear  quadratic form $Q_H(\chi)$ in   (\ref{eq:non-linerQ}) \cite{chan2018spectral}.

\subsection{Key Phases in HyperEF}
 As illustrated in Figure \ref{fig:overview}, HyperEF computes a much smaller hypergraph $H' = (V', E', w')$ given the original hypergraph $H = (V, E, w)$ by exploiting hyperedge effective resistances, where $|V'| < |V|$, $|E'| < |E|$, and $|w'|<|w|$. Specifically, HyperEF consists of the following three phases:  \textbf{Phase (A)}   constructs the Krylov subspace to approximate the eigensubspace related to the original hypergraph; \textbf{Phase (B)}    estimates the effective resistance of each hyperedge by applying the proposed optimization-based method; \textbf{Phase (C)}   constructs the coarsened hypergraph by aggregating node clusters with low effective-resistance diameters. For the sake of simplicity, key symbols used in this paper are summarized in Table \ref{table:symbols}. 
\subsection{Low-Resistance-Diameter Decomposition}
\begin{lemma}\label{lemma_bound}
Let $G = (\mathcal{V}, \mathcal{E}, z)$ be a weighted undirected graph with weights $z \in \mathbb{R}^\mathcal{V}_{> 0}$, sufficiently large $\gamma > 1$,   and the effective-resistance diameter be defined as $\max\limits_{u,v \in \mathcal{V}}R_{eff}(u,v)$. It is always possible to decompose a simple graph $G$ into multiple node clusters $G[\mathcal{V}_i]$ with low effective-resistance diameters  by removing only a constant fraction of edges \cite{alev2018graph}:
\begin{equation}
    \max\limits_{u,v \in \mathcal{V}_i} R_{eff_{G[\mathcal{V}_i]}}(u,v) \lesssim \gamma^3 \frac{|\mathcal{V}|}{z(\mathcal{E})}.
\end{equation}
\end{lemma}

\begin{lemma}\label{lemma_conductance}
Let $G = (\mathcal{V}, \mathcal{E}, z)$ be a weighted undirected graph with weights $z \in \mathbb{R}^\mathcal{V}_{\geq 0}$ and $\mathcal{C}_G$ be the conductance of $G$. The Cheeger's inequality allows obtaining   the following relationship between the effective-resistance diameter and the graph conductance \cite{alev2018graph}:
\begin{equation}
     \max\limits_{u,v \in \mathcal{V}}R_{eff}(u,v) \lesssim \frac{1}{\mathcal{C}_G^2}.
\end{equation}
   
\end{lemma}

The proposed HyperEF algorithm is based on extending the above theorems to hypergraph settings: Lemma \ref{lemma_bound} implies that we can decompose a hypergraph into multiple (hyperedge) clusters that have  small effective-resistance diameters by removing only a few inter-cluster hyperedges, while Lemma \ref{lemma_conductance} implies that   contracting the hyperedges (node clusters) with small effective-resistance diameters will not significantly impact the original hypergraph conductance.


\subsection{\textbf{Phase (A)}: Spectral Approximation via Krylov Subspace}
Based on Eq. (\ref{eq:non-linerQ}) and the optimization task in Eq. (\ref{eq:eff_resist}), we propose to estimate  effective resistance of a hyperedge $e$ by searching for a vector $\chi^*$ that  maximizes the following ratio: 
\begin{equation} \label{eq:eff_resist_hyper}
        R_e(\chi^*) =\max\limits_{\chi \in \mathbb{R}^\mathcal{V}} \frac{(\chi^{\top} b_{pq})^2}{Q_H(\chi)},  \quad p,q\in e
\end{equation}
To achieve high efficiency, our search for $\chi*$ will be limited to the eigensubspace spanned by a few Laplacian eigenvectors of the simple graph converted from the original hypergraph. Let $G_b = (\mathcal{V}_b, \mathcal{E}_b, z_b)$ be the bipartite graph corresponding to the hypergraph $H = (V, E, w)$, where $|\mathcal{V}_b| = |V| + |E|$, $|\mathcal{E}_b| = \Sigma_{e \in E}|e|$, and $z_b$ is the scaled edge weights: $z(e, p) = \frac{w(e)}{d(e)}$.



Eq. (\ref{eq:eff_resist0}) implies that  the approximate effective resistance of  each hyperedge can be obtained by  finding a few orthogonal eigenvectors that  maximize Eq. (\ref{eq:eff_resist_hyper}). 
To avoid high complexity of computing eigenvalues/eigenvectors,  we will leverage a scalable algorithm for approximating the eigenvectors by exploiting Krylov subspace defined as follows.
\begin{definition}
Given a nonsingular matrix $A_{n \times n}$, a vector $\chi \neq 0 \in \mathbb{R}^n$, the  order-$(\rho+1)$ Krylov subspace generated by $A$ from $x$ is
\begin{equation}\label{eq:krylov}
    \kappa_{\rho}(A, x) := span(x, Ax, A^2x, ..., A^{\rho}x),
\end{equation}
\end{definition}
where $x$ denotes a random vector, and $A$ denotes the normalized adjacency matrix. In our algorithm,  $A$ is obtained based on the simple graph converted from the hypergraph using star expansion.


\subsection{\textbf{Phase (B)}: Effective Resistance Estimation}
Assume that $x^{(1)}, x^{(2)}, ..., x^{(\rho)} \in \mathbb{R}^{\mathcal{V}_b}$ are the $\rho$ mutually-orthogonal vectors based on the order-$(\rho+1)$ Krylov subspace $\kappa_{\rho}(A, x)$ constructed in \textbf{Phase(A)}.  Our algorithm extends the effective resistance estimation method in (\ref{eq:eff_resist}) by incorporating the non-linear quadratic operator of hypergraphs (\ref{eq:non-linerQ}) to include the hypergraph spectral properties. By excluding the node embedding values associated with the star nodes in $x^{(i)}$, we generate a new set of vectors $\chi^{(i)}$ that are all mutually orthogonal. Next, each node in the hypergraph can be embedded into a $\rho$-dimensional space. Then, for each hyperedge $e$ its resistance ratio ($r_e$) associated with a   vector $\chi^{(i)} \in \chi^{(1)}, ..., \chi^{(\rho)}$ can be computed by
\begin{equation}\label{eq:ratios}
     r_e(\chi^{(i)}) = \frac{(\chi^{(i)\top} b_{pq})^2}{Q_H(\chi^{(i)})},  \quad p,q\in e
\end{equation}
where $p$ and $q$ are the two maximally-separated nodes in the $\rho$-dimensional embedding space. Eq. (\ref{eq:eff_resist_hyper}) returns multiple resistance ratios $r_e^{(1)},..., r_e^{(\rho)}$ corresponding to $\chi^{(1)},...,\chi^{(\rho)}$. 
After sorting resistance ratios in a descending order, we have
\begin{equation}
    r_e^1 > r_e^2 > ... > r_e^\rho.
\end{equation}
 In HyperEF, we chooses the top $m$ resistance ratios to estimate the effective resistance of each hyperedge. Specifically, we approximate the hyperedge effective resistance ($R_e$) by
\begin{equation}\label{eq:R}
    R_e = \sum\limits_{i=1}^{m} r_e^i, \quad e \in E.
\end{equation}
Note that for each hypergraph, the   Krylov subspace vectors in Eq. (\ref{eq:krylov}) only need to be computed once, which can be achieved in nearly linear time   using  only sparse matrix-vector operations.  The effective resistance of each hyperedge can be estimated  in constant time by identifying a few    ($m$) Krylov subspace vectors that   maximize the  resistance ratio in Eq. (\ref{eq:ratios}). 

\begin{figure}
    \centering
    \includegraphics [width = \linewidth]{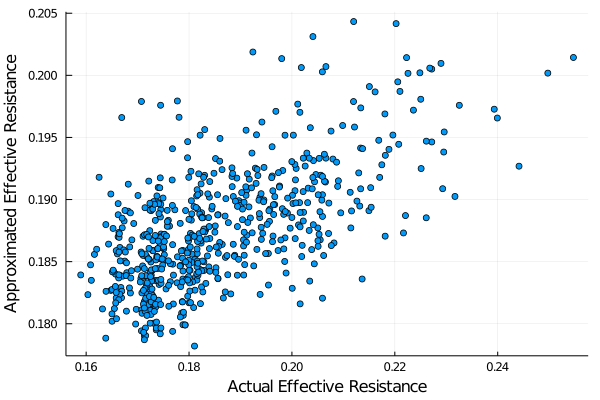}
    \caption{Actual effective resistance vs approximated effective resistance using our method for a   simple graph.}
    \label{fig:cor_fig}
\end{figure}
Since it is not clear how to efficiently compute the accurate hyperedge effective resistances for hypergraphs, we   only evaluate the proposed  method for a real-world simple graph. As shown in Figure \ref{fig:cor_fig},    the approximate edge effective resistances obtained using our method can well correlate with the ground truths. Algorithm \ref{alg:effR} provides the detailed flow of the proposed hyperedge effective resistance estimation method.

\begin{algorithm}
\small { \caption{The effective resistance estimation algorithm flow}\label{alg:effR}}
\textbf{Input:} Hypergraph $H = (V,E, w)$, $\rho$.\\
\textbf{Output:} {A vector of effective resistance $R$ with the size $|E|$}.\\
  \algsetup{indent=1em, linenosize=\small} \algsetup{indent=1em}
    \begin{algorithmic}[1]
    \STATE Construct the bipartite graph $G_b$ corresponding to $H$. 
    \STATE Construct the order-$(\rho+1)$ Krylov subspace.
    \STATE Use Gram–Schmidt method to obtain the  orthogonal vectors.
    \STATE For each hyperedge compute its $\rho$ resistance ratios  using (\ref{eq:ratios}).
    \STATE Obtain all hyperedge effective resistances $R$ based on (\ref{eq:R}).
     \STATE Return  $R$. 
    \end{algorithmic}
\end{algorithm}

\subsection{\textbf{Phase (C)}: Effective-Resistance Clustering}
 Lemma \ref{lemma_bound} implies that by removing only a few edges   it is possible to  decompose a simple graph into multiple node clusters with respect to the   effective-resistance diameter of   each cluster.  For hypergraphs, HyperEF adopts the similar idea for identifying node clusters. 
 
 Specifically, for a given hypergraph $H = (V, E, w)$,  assume that $R \in \mathbb{R}_{>0}^{E}$ is a vector including all hyperedge effective resistances   computed in \textbf{Phase(B)}. Then,  HyperEF will first exploit the effective resistance vector $R$   to decompose a hypergraph into multiple node clusters with relatively low effective-resistance diameters. Specifically, HyperEF will produce   node clusters such that the     effective-resistance diameter of each cluster is strictly lower than a given threshold $\delta \in \mathbb{R}_{>0}$. Subsequently, by treating each node cluster as a new node, HyperEF will create  a much smaller hypergraph $H' = (V', E', w')$  that can well preserve the spectral (structural) properties of the original hypergraph $H$.   Since the nodes within each cluster are strongly   coupled with each other, contracting such clusters (or hyperedges)  will not significantly alter the structural (spectral) properties of the original hypergraph.

In our implementation, HyperEF will contract the entire hyperedge $e$ with low effective resistance (below a given threshold $\delta$) if all the nodes within $e$ have not been processed (clustered) before. Otherwise, HyperEF will only contract the unprocessed nodes within the hyperedge.

\section{Multilevel Hypergraph Coarsening}\label{sec:multilevel}
In this section, we extend  the proposed spectral hypergraph coarsening method   to  a multilevel coarsening framework that iteratively computes a vector of effective resistance $R$ and contract the hyperedges with low effective resistances ($R_e < \delta$). Our algorithm iteratively contracts the node clusters by merging the nodes within each cluster and replacing that cluster with a new node that we call \textit{supernode} in the next level. We transfer the hypergraph structural information through the levels by assigning a  weight (that is equal to the hyperedge effective resistance evaluated at the previous level) to the   supernode corresponding to each cluster. 

\begin{definition}
Let $H^{(l)} = (V^{(l)}, E^{(l)}, w^{(l)})$ be the hypergraph at the $l$-th level,    $S \in V^{(l)}$ be a cluster of nodes  that produces a supernode $\vartheta \in V^{(l+1)}$ at the level   $l+1$. We define the vector of node weights to be $\eta^{(l)} \in \mathbb{R}_{\geq 0}^{V^{(l)}}$ and initialize it with all zeros for the original hypergraph. We update $\eta$ at   level $l$ by
\begin{equation}\label{eq:Nweights}
    \eta_\vartheta := \sum\limits_{j = 1}^{|S|} \eta(v_j^{(l)}). \quad v\in V^{(l)}: v\in S
\end{equation}
\end{definition}

\textbf{Node Weight Propagation (NWP).} The vector of effective resistance $R$ will be updated according to the node weights $\eta$ to pass the clustering information obtained from previous levels to the current level:
\begin{equation}\label{eq:R_W_update}
     R_e^{(l)} = \sum\limits_{k=1}^{|e|}\eta(v_k^{(l)}) + R_e^{(l)}.
\end{equation}
As the result, the hyperedge effective resistance at a coarse level not only depends on the evaluated effective resistance ($R_e^{(l)}$) computed at the current level but also   the results   transferred from all the previous (finer) levels. 

As observed in our extensive experiment results,   using (\ref{eq:R_W_update}) for estimating effective resistances  leads to more   balanced hypergraph clustering results when compared with the implementation that ignores the previous clustering information.   The complete HyperEF algorithm flow has been shown in Algorithm \ref{alg:HyperEF}  for  coarsening a given hypergraph $H$ within $L$ levels. 

\begin{algorithm}
\small { \caption{HyperEF algorithm flow}\label{alg:HyperEF}}
\textbf{Input:} Hypergraph $H = (V,E, w)$, $\delta$, $L$, $\eta$.\\
\textbf{Output:} {A coarsened hypergraph $H' = (V', E', w')$ that $|V'| \ll |V|$}.\\
  \algsetup{indent=1em, linenosize=\small} \algsetup{indent=1em}
    \begin{algorithmic}[1]
    \STATE Initialize $H'$ $\leftarrow$ $H$
    \FOR{$l \gets 1$ to $L$}
    \STATE Call Algorithm \ref{alg:effR} to compute a vector of effective resistance $R$ with the size $|E'|$ for given hypergraph $H'$.
    \STATE Compute the node weights using (\ref{eq:Nweights}).
    \STATE Update the effective resistance vector $R$ by applying (\ref{eq:R_W_update}).
    \STATE Sort the hyperedges with ascending $R$ values.
    \STATE  Starting with the hyperedges that have the lowest effective resistances, contract (cluster) the hyperedge (nodes) if $R_e< \delta$.
    \STATE Construct a coarsened hypergraph $H'$ accordingly.
    \ENDFOR
    \STATE Return $H'$.
     
    \end{algorithmic}
\end{algorithm}

\section{Complexity Analysis}\label{sec:complexity_analysis}
The algorithm complexity of \textbf{Phase(A)} for constructing the Krylov subspace for the bipartite graph $G_b = (\mathcal{V}_b, \mathcal{E}_b, z_b)$ corresponding to the original hypergraph $H= (V, E, w)$ is $O(|\mathcal{E}_b|)$; the complexity of the hyperedge effective resistance estimation and hyperedge clustering corresponding to \textbf{Phase(B)} and \textbf{Phase(C)} is $O(\rho|E|)$; the complexity of computing the node weights through the multilevel framework is $O(|E|)$ that leads to the overall nearly-linear algorithm complexity of $O(\rho|E|+|\mathcal{E}_b|)$.

\section{Experimental Results}\label{sec:results}
This section presents the results of a variety of experiments for evaluating the performance and efficiency of the proposed spectral hypergraph coarsening algorithm (HyperEF). The real-world VLSI design benchmarks ``ibm01'', ``ibm02'', ..., ``ibm18'' with   $13,000$ to $210,000$ cells have been adopted  \footnote[1]{https://vlsicad.ucsd.edu/UCLAWeb/cheese/ispd98.html}. All experiments have been evaluated on a laptop with 8 GB of RAM and a 2.2 GHz Quad-Core Intel Core i7 processor.

\subsection{Experiment Setup}
  The HyperEF algorithm has been implemented as follows: (1) construct the bipartite graph $G_b$ corresponding to the hypergraph $H$ (2) generate a random vector $x$ orthogonal to the all-one vector \textbf{1} satisfying $1^\top x = 0$; (3) construct a $\rho$-order Krylov subspace using $x$ where $\rho=200$; (4) select $10$ vectors from the Krylov subspace for embedding each node   into a $10$-dimensional space; (5) compute $10$   hyperedge resistance ratios; (6) the estimated effective resistance of each hyperedge is based on the top resistance ratio ($m$ = 1);
  (7) $\delta$ is set to be the maximum hyperedge effective resistance at each level. The input arguments of hMetis are set as default values and UBfactor $ =5$ through all the experiments. An implementation of our algorithm and the code for reproducing our experimental results are available online at \ULurl{https://github.com/Feng-Research/HyperEF}.
  

\subsection{Node Weight Propagation (NWP)}
The proposed node weight propagation technique preserves the spectral features by passing the effective resistance values along the levels.  It allows us to exploit the previous cluster information for computing hyperedge effective resistances of the current level. To illustrate the benefit of NWP,
we conduct an experiment for a small   hypergraph $H$ illustrated in Figure \ref{fig:overview}, to compare the estimated effective resistances with and without employing the NWP scheme. We apply HyperEF for $L = 2$ levels and contract one hyperedge at each level to create a coarsened hypergraph with $6$ nodes and $2$ hyperedges. Table \ref{tab:example} summarizes the estimated effective resistance of every hyperedge   with and without using the NWP method at different levels. We observe that the employing NWP technique leads to more accurate estimation results in both levels.

\begin{table}[]
\centering
\caption{Impact of the node weight propagation (NWP) technique on estimating effective resistances.}
\begin{tabular}{|c|ccc|cc|}
\hline
 & \multicolumn{3}{c|}{level 1} & \multicolumn{2}{c|}{level 2} \\ \hline
            & \multicolumn{1}{c|}{$R_1$}  & \multicolumn{1}{c|}{$R_2$}  & $R_3$   & \multicolumn{1}{c|}{$R_1$}  & $R_2$  \\ \hline
without NWP & \multicolumn{1}{c|}{2.5} & \multicolumn{1}{c|}{3.3} & 1.8  & \multicolumn{1}{c|}{3.1} & 2.8 \\ \hline
with NWP    & \multicolumn{1}{c|}{2.6} & \multicolumn{1}{c|}{4.2} & 1.96 & \multicolumn{1}{c|}{5.1} & 6.1 \\ \hline
\end{tabular}
\label{tab:example}
\end{table}

\subsection{Spectral Hypergraph  Decomposition}
In this section, we provide comprehensive experimental results to evaluate the performance of the proposed spectral hypergraph coarsening  algorithm by comparing HyperEF with both spectral and non-spectral   coarsening methods. We leverage HyperEF to decompose the hypergraph into multiple clusters by repeatedly clustering the nodes and aggregating the vertices within each cluster. The {Cheeger's inequality} implies that the quality of a cluster $S$ is higher if the conductance of $S$ is smaller. Accordingly, we utilize the following average conductance of clusters as a measure to analyze the performance of each method
\begin{equation}
    \mathcal{C} = \frac{1}{N}\sum\limits_{i=1}^N \mathcal{C}(S_i),
\end{equation}
where $N$ is the number of the clusters and $\mathcal{C}(S_i)$ is the   conductance of a node cluster $S_i$. Additionally, our algorithm can accept a set of node weights proportional to the size of the cells that is beneficial for VLSI applications. In this case, our objective denominator in \ref{def1} will be modified by adding the node weights to the node degrees.

\begin{figure}
    \centering
    \includegraphics [width =\columnwidth]{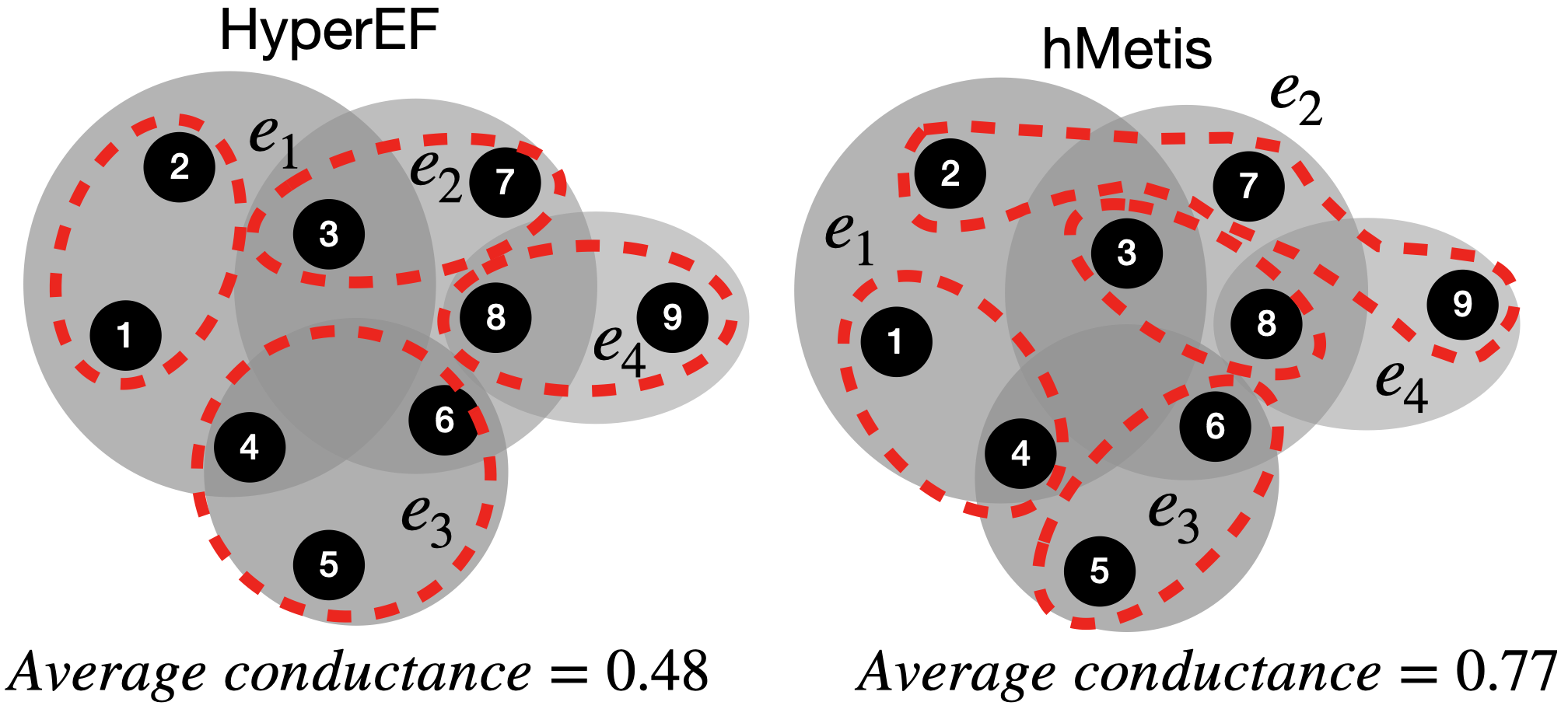}
    \caption{Hypergraph clustering results: HyperEF vs hMetis.}
    \label{fig:sample_case}
\end{figure}

We compare HyperEF with the SoTA hypergraph partitioner, hMetis, in terms of performance and runtime. HyperEf is compared with the hMetis partitioning results by considering the hypergraph conductance metric (\ref{def1}). It should be noted that a direct comparison of our proposed coarsening algorithm with the coarsening method used in hMetis is impossible since the code is unavailable.

Figure \ref{fig:sample_case} demonstrates the node clustering results obtained using HyperEF and hMetis.  Both methods partition the hypergraph into 4 clusters, and the average conductance of clusters has been computed to evaluate the performance of each method. HyperEF computes the effective resistance of hyperedges, and sort them accordingly: $R_{e_4} < R_{e_3} < R_{e_2} < R_{e_1}$. In this example, $e_4$ has the smallest effective resistance resulting in clustering nodes 8 and 9. The next hyperedge with the smallest effective resistance is $e_3$, forming another cluster including nodes 4, 5, and 6. Then, $e_2$ has been discovered, and nodes 3 and 7 are clustered together since nodes 6 and 8 have already been processed. Finally, HyperEF explores $e_1$ and produces another node cluster containing nodes 1 and 2, since nodes 3 and 4 have been previously clustered. The results show that HyperEF outperforms hMetis by creating clusters with a significantly lower average conductance.

\begin{figure}
    \centering
    \includegraphics [width = \linewidth]{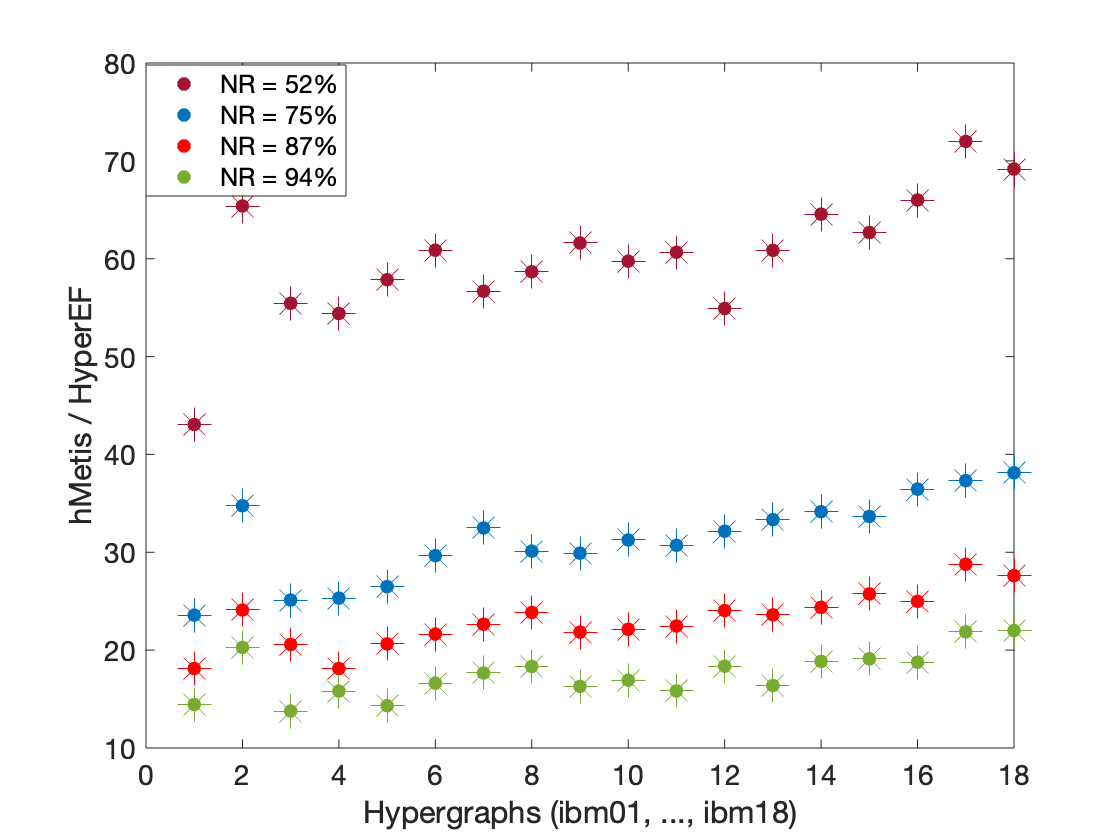}
    \caption{Comparison of runtime performance.}
    \label{fig:runtime}
\end{figure}

Table \ref{nr52}, Table \ref{nr75}, Table \ref{nr87}, and Table \ref{nr94} show the average conductance of clusters $\mathcal{C}$ computed with both HyperEF and hMetis by decomposing the hypergraph into the same number of node clusters with the same node reduction ratios (NRs) for ``ibm01'', ..., ``ibm18'' datasets. HyperEF achieves NR = $52\%$\footnote{If an original hypergraph has 100 nodes, the coarsened hypergraph will have 48 nodes.} with $L$=1, NR = $75\%$ with $L$=2, NR = $87\%$ with $L$=3, and NR = $94\%$ with $L$=4. All the NRs are chosen unbiased to evaluate the performance of HyperEF for different coarsening ratios. Our extensive results show that HyperEF always produces better results (lower average conductance) for all the test cases with NRs from $52\%$ to $87\%$, and similar results with NR = $94\%$. We also report the runtime $\mathcal{T}$ (seconds) of both methods for decomposing the hypergraphs. Figure \ref{fig:runtime} shows that HyperEF achieves up to $72 \times$ speedup over hMetis.

\begin{figure}
    \centering
    \includegraphics [width = \linewidth]{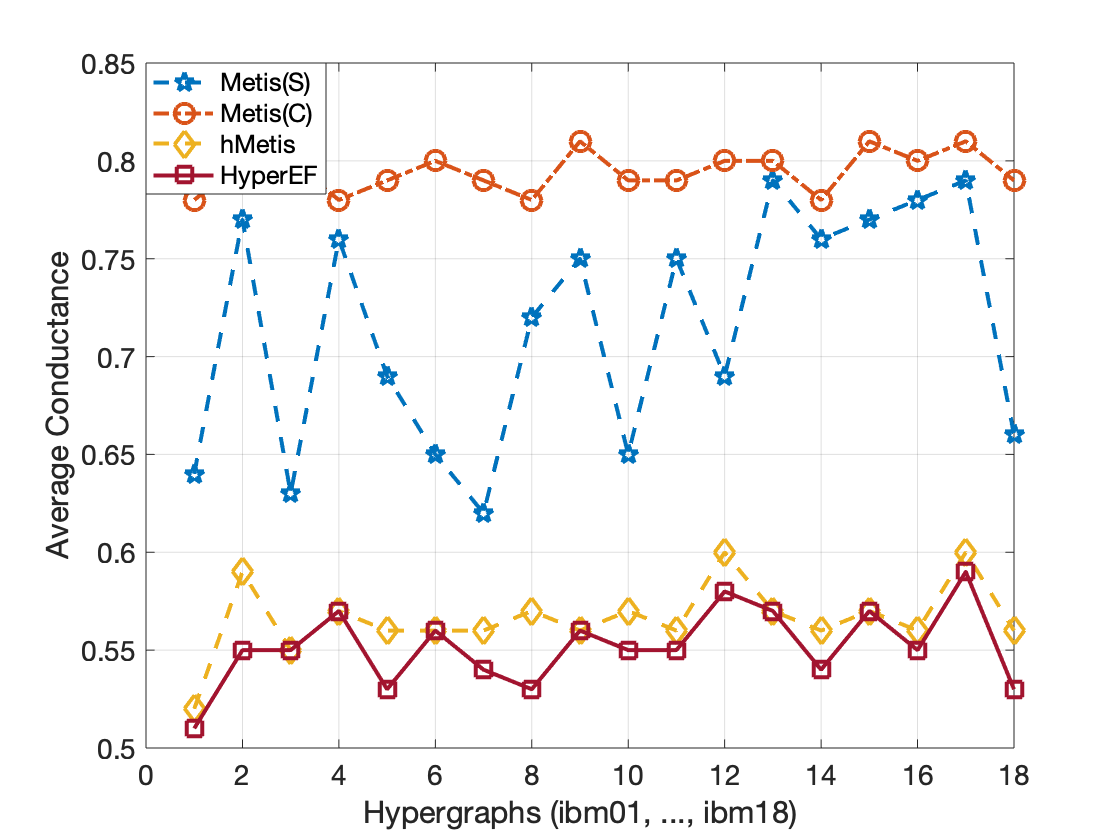}
    \caption{HyperEF vs non-spectral methods.}
    \label{fig:nonspectral}
\end{figure}

To further investigate the performance of HyperEF, we designed various experiments to decompose the hypergraph using spectral and non-spectral methods. For spectral hypergraph coarsening benchmarks, traditional simple graph spectral coarsening techniques are leveraged to decompose   hypergraphs \cite{zhao2021towards} after converting them  into simple graphs using star and clique expansions. Additionally, we compare HyperEF with HyperSF, which is a recently developed spectral hypergraph coarsening method \cite{HyperSF}. Note that in \cite{HyperSF},  the average local conductance values are reported, whereas, in our experiments,   the average global conductance of clusters is reported. For all the above experiments, we decompose   hypergraphs into the same number of node clusters with NR = $83\%$. 
\begin{figure}
    \centering
    \includegraphics [width = \linewidth]{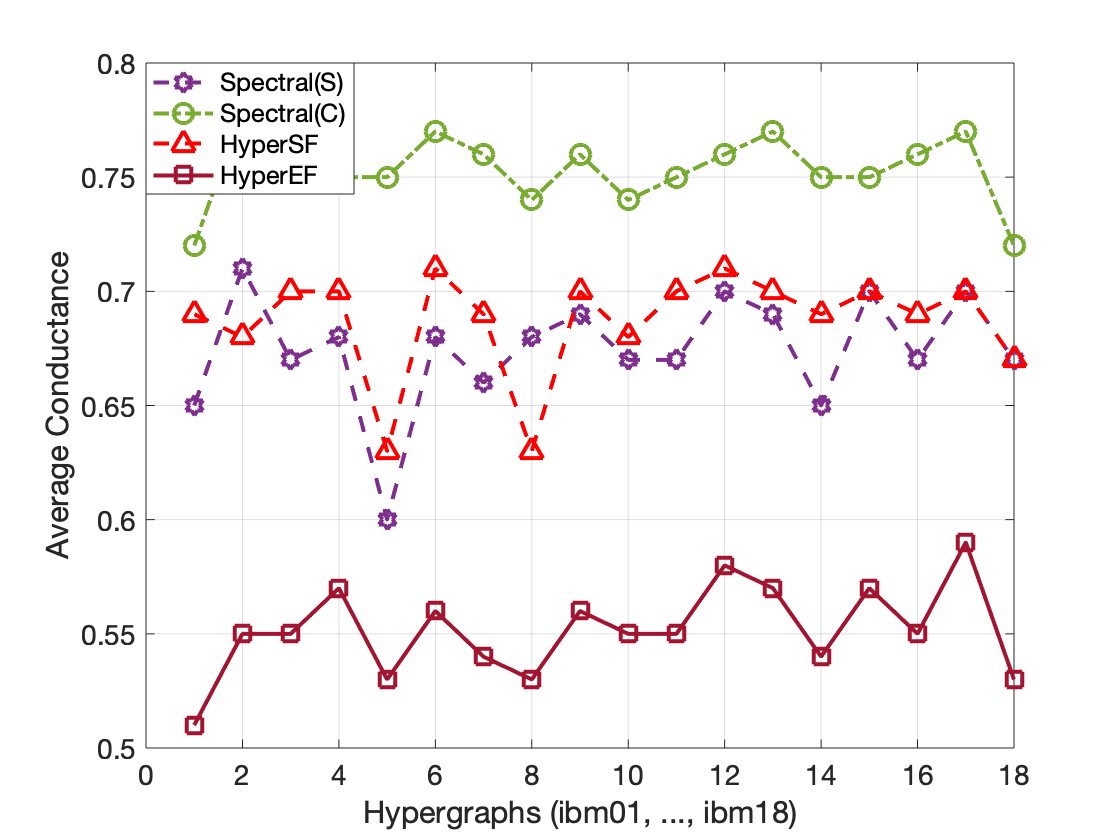}
    \caption{HyperEF vs spectral methods.}
    \label{fig:spectral}
\end{figure}
Figure \ref{fig:spectral} shows that HyperEF beats all other spectral coarsening methods by achieving the lowest conductance for all the test cases.
For non-spectral methods, we use Metis to partition the simple graph corresponding to the hypergraph (achieved by applying star and clique expansions). In Figure \ref{fig:nonspectral}, we compare the performance of Metis (star and clique expansions), hMetis and HyperEF. We observe that HyperEF outperforms all non-spectral methods by returning the lowest average conductance when NR = $83\%$.


\begin{table}[]
\caption{NR = $52\%$, and $L$=1 (HyperEF only).}
\label{nr52}
\scriptsize
\centering
\begin{tabular}{|c|c|c|c|c|c|}
\hline
ibm & $N_{cluster}$     & $\mathcal{C}$(HyperEF)                  & $\mathcal{C}$(hMetis) & $\mathcal{T}$(HyperEF)                  & $\mathcal{T}$(hMetis) \\ \hline
1   & 6,183  & {\textbf{0.75}} & 0.81      & {\textbf{0.72}} & 31 (\textbf{43$\times$})   \\ \hline
2   & 8,746  & {\textbf{0.74}} & 0.8       & {\textbf{0.78}} & 51 (\textbf{65$\times$})   \\ \hline
3   & 10,755 & {\textbf{0.76}} & 0.8       & {\textbf{1.01}} & 56 (\textbf{55$\times$})   \\ \hline
4   & 12,713 & {\textbf{0.76}} & 0.81      & {\textbf{1.25}} & 68 (\textbf{54$\times$})   \\ \hline
5   & 12,334 & {\textbf{0.69}} & 0.75      & {\textbf{1.21}} & 70 (\textbf{58$\times$})   \\ \hline
6   & 14,935 & {\textbf{0.77}} & 0.81      & {\textbf{1.38}} & 84 (\textbf{61$\times$})   \\ \hline
7   & 21,727 & {\textbf{0.77}} & 0.81      & {\textbf{2.17}} & 123 (\textbf{57$\times$})  \\ \hline
8   & 23,285 & {\textbf{0.75}} & 0.81      & {\textbf{2.25}} & 132 (\textbf{59$\times$})  \\ \hline
9   & 23,888 & {\textbf{0.76}} & 0.81      & {\textbf{2.24}} & 138 (\textbf{62$\times$})  \\ \hline
10  & 32,427 & {\textbf{0.76}} & 0.8       & {\textbf{3.23}} & 193 (\textbf{60$\times$})  \\ \hline
11  & 31,443 & {\textbf{0.77}} & 0.81      & {\textbf{3.05}} & 185 (\textbf{61$\times$})  \\ \hline
12  & 32,530 & {\textbf{0.78}} & 0.82      & {\textbf{3.66}} & 201 (\textbf{55$\times$})  \\ \hline
13  & 39,473 & {\textbf{0.78}} & 0.82      & {\textbf{3.96}} & 241 (\textbf{61$\times$})  \\ \hline
14  & 68,607 & {\textbf{0.75}} & 0.8       & {\textbf{6.32}} & 408 (\textbf{65$\times$})  \\ \hline
15  & 72192  & {\textbf{0.78}} & 0.82      & {\textbf{8.01}} & 502 (\textbf{63$\times$})  \\ \hline
16  & 81,431 & {\textbf{0.76}} & 0.81      & {\textbf{8.44}} & 557 (\textbf{66$\times$})  \\ \hline
17  & 81,789 & {\textbf{0.78}} & 0.83      & {\textbf{8.46}} & 609 (\textbf{72$\times$})  \\ \hline
18  & 93,000 & {\textbf{0.74}} & 0.81      & {\textbf{9.34}} & 646 (\textbf{69$\times$})  \\ \hline
\end{tabular}
\end{table}

\begin{table}[]
\caption{NR = $75\%$, and $L$=2 (HyperEF only).}
\label{nr75}
\scriptsize
\centering
\begin{tabular}{|c|c|c|c|c|c|}
\hline
ibm & $N_{cluster}$     & $\mathcal{C}$(HyperEF)                  & $\mathcal{C}$(hMetis) & $\mathcal{T}$(HyperEF)                   & $\mathcal{T}$(hMetis) \\ \hline
1   & 3,160  & {\textbf{0.62}} & 0.65      & {\textbf{1.23}}  & 29 (\textbf{24$\times$})   \\ \hline
2   & 4,314  & {\textbf{0.62}} & 0.67      & {\textbf{1.41}}  & 49 (\textbf{35$\times$})   \\ \hline
3   & 5,329  & {\textbf{0.63}} & 0.66      & {\textbf{2.11}}  & 53 (\textbf{25$\times$})   \\ \hline
4   & 6,300  & {\textbf{0.64}} & 0.66      & {\textbf{2.37}}  & 60 (\textbf{25$\times$})   \\ \hline
5   & 6,140  & {\textbf{0.59}} & 0.63      & {\textbf{2.34}}  & 62 (\textbf{26$\times$})   \\ \hline
6   & 7,353  & {\textbf{0.64}} & 0.66      & {\textbf{2.63}}  & 78 (\textbf{30$\times$})   \\ \hline
7   & 10,900 & {\textbf{0.63}} & 0.67      & {\textbf{3.54}}  & 115 (\textbf{32$\times$})  \\ \hline
8   & 11,800 & {\textbf{0.61}} & 0.67      & {\textbf{4.15}}  & 125 (\textbf{30$\times$})  \\ \hline
9   & 11,362 & {\textbf{0.64}} & 0.66      & {\textbf{4.38}}  & 131 (\textbf{30$\times$})  \\ \hline
10  & 16,052 & {\textbf{0.63}} & 0.67      & {\textbf{5.79}}  & 181 (\textbf{31$\times$})  \\ \hline
11  & 16,070 & {\textbf{0.64}} & 0.67      & {\textbf{5.73}}  & 176 (\textbf{31$\times$})  \\ \hline
12  & 16,207 & {\textbf{0.65}} & 0.7       & {\textbf{5.94}}  & 191 (\textbf{32$\times$})  \\ \hline
13  & 19,617 & {\textbf{0.65}} & 0.68      & {\textbf{6.87}} & 229 (\textbf{33$\times$})  \\ \hline
14  & 34,136 & {\textbf{0.62}} & 0.66      & {\textbf{11.51}} & 393 (\textbf{34$\times$})  \\ \hline
15  & 35,524 & {\textbf{0.66}} & 0.69      & {\textbf{14.44}} & 486 (\textbf{34$\times$})  \\ \hline
16  & 39,645 & {\textbf{0.63}} & 0.67      & {\textbf{14.62}} & 533 (\textbf{36$\times$})  \\ \hline
17  & 39,951 & {\textbf{0.66}} & 0.7       & {\textbf{15.22}} & 568 (\textbf{37$\times$})  \\ \hline
18  & 47,702 & {\textbf{0.6}}  & 0.67      & {\textbf{15.79}} & 602 (\textbf{38$\times$})  \\ \hline
\end{tabular}
\end{table}

\begin{table}[]
\caption{NR = $87\%$, and $L$=3 (HyperEF only).}
\label{nr87}
\scriptsize
\centering
\begin{tabular}{|c|c|c|c|c|c|}
\hline
ibm & $N_{cluster}$     & $\mathcal{C}$(HyperEF)                  & $\mathcal{C}$(hMetis)                   & $\mathcal{T}$(HyperEF)                   & $\mathcal{T}$(hMetis) \\ \hline
1   & 1,642  & {\textbf{0.51}} & 0.52                        & {\textbf{1.49}}  & 27 (\textbf{18$\times$})   \\ \hline
2   & 2,342  & {\textbf{0.55}} & 0.59                        & {\textbf{1.95}}  & 47 (\textbf{24$\times$})   \\ \hline
3   & 2,712  & {\textbf{0.55}} & {\textbf{0.55}} & {\textbf{2.43}}  & 50 (\textbf{21$\times$})   \\ \hline
4   & 3,311  & {\textbf{0.57}} & {\textbf{0.57}} & {\textbf{2.98}}  & 54 (\textbf{18$\times$})   \\ \hline
5   & 3,128  & {\textbf{0.53}} & 0.56                        & {\textbf{2.81}}  & 58 (\textbf{21$\times$})   \\ \hline
6   & 3,713  & {\textbf{0.56}} & {\textbf{0.56}} & {\textbf{3.42}}  & 74 (\textbf{22$\times$})   \\ \hline
7   & 5,501  & {\textbf{0.54}} & 0.56                        & {\textbf{4.68}}  & 106 (\textbf{23$\times$})  \\ \hline
8   & 6,369  & {\textbf{0.53}} & 0.57                        & {\textbf{5.03}}  & 120 (\textbf{24$\times$})  \\ \hline
9  & 5,798  & {\textbf{0.56}} & {\textbf{0.56}} & {\textbf{5.63}}  & 123 (\textbf{22$\times$}) \\ \hline
10  & 8,400  & {\textbf{0.55}} & 0.57                        & {\textbf{7.23}}  & 160 (\textbf{22$\times$})  \\ \hline
11  & 8,371  & {\textbf{0.55}} & 0.56                        & {\textbf{7.39}}  & 166 (\textbf{22$\times$})  \\ \hline
12  & 8,458  & {\textbf{0.58}} & 0.6                         & {\textbf{7.57}}  & 182 (\textbf{24$\times$})  \\ \hline
13 & 10,026 & {\textbf{0.57}} & {\textbf{0.57}} & {\textbf{9.15}}  & 216 (\textbf{24$\times$}) \\ \hline
14  & 17,509 & {\textbf{0.54}} & 0.56                        & {\textbf{14.77}} & 360 (\textbf{24$\times$})  \\ \hline
15 & 18,070 & {\textbf{0.57}} & {\textbf{0.57}} & {\textbf{17.79}} & 458 (\textbf{26$\times$}) \\ \hline
16  & 19,992 & {\textbf{0.55}} & 0.56                        & {\textbf{19.61}} & 490 (\textbf{25$\times$})  \\ \hline
17  & 20,551 & {\textbf{0.59}} & 0.6                         & {\textbf{19.46}} & 560 (\textbf{29$\times$})  \\ \hline
18  & 25,315 & {\textbf{0.53}} & 0.56                        & {\textbf{21.04}} & 581 (\textbf{28$\times$})  \\ \hline
\end{tabular}
\end{table}

\begin{table}[]
\centering
\caption{NR = $94\%$, and $L$=4 (HyperEF only).}
\label{nr94}
\scriptsize
\begin{tabular}{|c|c|c|c|c|c|}
\hline
ibm & $N_{cluster}$     & $\mathcal{C}$(HyperEF) & $\mathcal{C}$(hMetis)                   & $\mathcal{T}$(HyperEF)                   & $\mathcal{T}$(hMetis) \\ \hline
1   & 862    & 0.45       & {\textbf{0.41}} & {\textbf{1.73}}  & 25 (\textbf{14$\times$})   \\ \hline
2   & 1,350  & 0.52       & {\textbf{0.53}} & {\textbf{2.17}}  & 44 (\textbf{20$\times$})   \\ \hline
3   & 1,395  & 0.49       & {\textbf{0.45}} & {\textbf{3.34}}  & 46 (\textbf{14$\times$})   \\ \hline
4   & 1,735  & 0.52       & {\textbf{0.46}} & {\textbf{3.35}}  & 53 (\textbf{16$\times$})   \\ \hline
5  & 1,619  & {\textbf{0.49}} & {\textbf{0.49}} & {\textbf{3.77}}  & 54 (\textbf{14$\times$})  \\ \hline
6   & 1,888  & 0.5        & {\textbf{0.46}} & {\textbf{3.97}}  & 66 (\textbf{17$\times$})   \\ \hline
7   & 2,836  & 0.48       & {\textbf{0.46}} & {\textbf{5.49}} & 97 (\textbf{18$\times$})   \\ \hline
8  & 3,574  & {\textbf{0.49}} & {\textbf{0.49}} & {\textbf{5.89}}  & 108 (\textbf{18$\times$}) \\ \hline
9   & 3,017  & 0.49       & {\textbf{0.45}} & {\textbf{6.87}} & 112 (\textbf{16$\times$})  \\ \hline
10  & 4,481  & 0.49       & {\textbf{0.48}} & {\textbf{9.45}}  & 160 (\textbf{17$\times$})  \\ \hline
11  & 4,391  & 0.5        & {\textbf{0.46}} & {\textbf{9.79}}  & 155 (\textbf{16$\times$})  \\ \hline
12  & 4,528  & 0.54       & {\textbf{0.52}} & {\textbf{9.04}}  & 166 (\textbf{18$\times$})  \\ \hline
13  & 5,174  & 0.52       & {\textbf{0.47}} & {\textbf{12.39}} & 203 (\textbf{16$\times$})  \\ \hline
14  & 9,055  & 0.48       & {\textbf{0.47}} & {\textbf{17.93}} & 338 (\textbf{19$\times$})  \\ \hline
15  & 9,277  & 0.51       & {\textbf{0.47}} & {\textbf{22.03}} & 421 (\textbf{19$\times$})  \\ \hline
16  & 10,308 & 0.49       & {\textbf{0.47}} & {\textbf{23.07}} & 433 (\textbf{19$\times$})  \\ \hline
17  & 10,782 & 0.54       & {\textbf{0.51}} & {\textbf{23.69}} & 518 (\textbf{22$\times$})  \\ \hline
18 & 13,864 & {\textbf{0.48}} & {\textbf{0.48}} & {\textbf{24.72}} & 544 (\textbf{22$\times$}) \\ \hline
\end{tabular}
\end{table}

\subsection{Cut (Conductance) Preservation in Coarsened Hypergraphs }
To further evaluate the performance of HyperEF, we incorporate HyperEF with hMetis to partition the hypergraphs and compare the cut and conductance values before and after spectral hypergraph coarsening. Initially, we coarsen the hypergraph using HyperEF to generate a smaller hypergraph   and then use hMetis to bipartite the coarsened hypergraph. This experiment compares the following two results: \textbf{Part 1:} bisect the original hypergraph using hMetis and compute the cut and conductance; \textbf{Part 2:} coarsen the original hypergraph using HyperEF, and subsequently use hMetis to bisect the coarsened hypergraph. For \textbf{Part 2}, all the node partitions associated with the coarsened hypergraphs are directly mapped to the original hypergraph nodes before the cut and conductance values are computed. Table \ref{tab:bisect} shows that the coarsened hypergraphs can reasonably retain the key spectral properties of the original hypergraph by well preserving the cut and conductance values. In Table \ref{tab:bisect}, $H$ is the original hypergraph (\textbf{Part 1}), and $H'$ is the coarsened hypergraph (\textbf{Part 2}).  NR and ER are the node reduction ratios and hyperedge reduction ratios, respectively. These results imply that HyperEF   only clusters the strongly-coupled nodes  and will not significantly impact the cut and conductance after coarsening, leaving the global hypergraph structure intact. 
\begin{table}
\centering
\scriptsize
\caption{The results of HyperEF for preserving the spectral hypergraph properties. }
\begin{tabular}{|c|c|c|c|c|c|c|}
\hline
ibm & NR(\%) & ER(\%) & cut(H) & cut(H') & $\mathcal{C}$(H)   & $\mathcal{C}$(H')  \\ \hline
01       & 31 & 25 & 180    & 185      & 0.0082 & 0.0085 \\ \hline
02       & 30 & 28 & 264    & 278      & 0.0069 & 0.0073 \\ \hline
03       & 31 & 28 & 954    & 996      & 0.022  & 0.022  \\ \hline
04       & 35 & 25 & 537    & 544      & 0.0108 & 0.0111 \\ \hline
05       & 32 & 29 & 1708   & 1700     & 0.0293 & 0.03   \\ \hline
06       & 34 & 28 & 899    & 906      & 0.0158 & 0.0162 \\ \hline
07       & 31 & 27 & 859    & 887      & 0.0109 & 0.0113 \\ \hline
08       & 29 & 28 & 1147   & 1165     & 0.0116 & 0.0121 \\ \hline
09       & 28 & 23 & 633    & 678      & 0.0057 & 0.0061 \\ \hline
10       & 34 & 28 & 1286   & 1379     & 0.01   & 0.0094 \\ \hline
11       & 31 & 23 & 962    & 1070     & 0.0073 & 0.0079 \\ \hline
12       & 31 & 26 & 1893   & 1975     & 0.0132 & 0.0139 \\ \hline
13       & 32 & 24 & 840    & 921      & 0.0049 & 0.0052 \\ \hline
14       & 28 & 26 & 1858   & 1959     & 0.0072 & 0.0076 \\ \hline
15       & 26 & 21 & 2670   & 2869     & 0.0085 & 0.0086 \\ \hline
16       & 31 & 27 & 1746   & 1865     & 0.0049 & 0.0052 \\ \hline
17       & 28 & 25 & 2222   & 2380     & 0.0055 & 0.0059 \\ \hline
18       & 27 & 26 & 1881   & 1947     & 0.0048 & 0.0049 \\ \hline
\end{tabular}
\label{tab:bisect}
\end{table}

\subsection{Runtime Scalability}
\begin{figure}
    \centering
    \includegraphics [width = 0.9\linewidth]{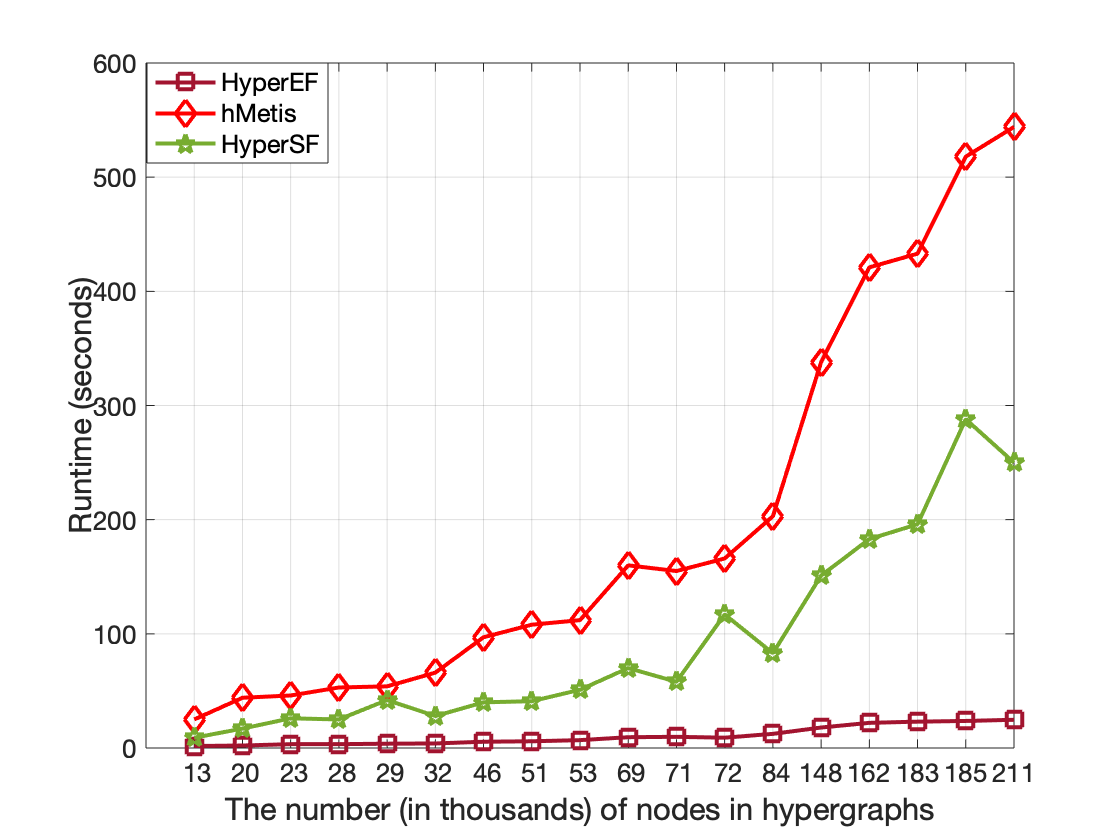}
    \caption{Runtime scalability comparisons.}
    \label{fig:scalability}
\end{figure}
Figure \ref{fig:scalability} shows that HyperEF has  much better  runtime scalability when compared with HyperSF and hMetis, achieving about $70\times$ speedups over hMetis and $20\times$ speedups over HyperSF.
\vspace{-10pt}
\section{Conclusion}\label{sec:conclusion}
This work presents a highly-efficient spectral hypergraph coarsening algorithm (HyperEF) that allows creating a much smaller hypergraph while preserving the key spectral (structural) properties of the original hypergraph. HyperEF is built upon a novel effective resistance estimation method for decomposing a hypergraph into multiple strongly-coupled node clusters. To achieve more effective decomposition, a node weight propagation  scheme  is  introduced to allow extending HyperEF to a multilevel  hypergraph coarsening framework.  When compared to state-of-the-art  methods, our extensive experiment results on real-world VLSI test cases show that HyperEF can significantly improve the hypergraph   clustering (partitioning) quality  while achieving up to $70\times$ runtime speedups.

\section{Acknowledgments}
This work is supported in part by  the National Science Foundation under Grants    CCF-2021309,  CCF-2011412, CCF-2212370, and CCF-2205572.

\bibliographystyle{abbrv}
\bibliography{imported_bib}

\end{document}